\pdfoutput=1

\documentclass[11pt]{article}

\usepackage[final]{acl}

\usepackage{times}
\usepackage{latexsym}

\usepackage[T1]{fontenc}

\usepackage[utf8]{inputenc}

\usepackage{microtype}

\usepackage{inconsolata}

\usepackage{graphicx}

\usepackage{hyperref}
\usepackage{url}

\usepackage[linesnumbered,ruled,vlined]{algorithm2e}

\usepackage{float}
\usepackage{amsmath}
\usepackage{amssymb}
\usepackage{booktabs}
\usepackage{makecell}
\usepackage{colortbl}
\usepackage{xcolor}

\usepackage{multicol}
\usepackage{multirow}
\usepackage{blindtext}
\usepackage{dirtree}
\usepackage{tikz}

\usepackage{amsthm}
\usepackage{pbox}
\usepackage{chngpage}
\usepackage{minitoc}
\usepackage{longtable}
\usepackage{comment}
\usepackage{tcolorbox}

\nolinenumbers 

\title{Path Pooling: Training-Free Structure Enhancement for Efficient Knowledge Graph Retrieval-Augmented Generation}
\author{
	  Hairu Wang$^{1,3}$\thanks{Equal contribution.}, Yuan Feng$^{1,3}$\footnotemark[1],
      Xike Xie$^{2,3}$\thanks{Corresponding author.}, S Kevin Zhou$^{2,3}$ \\
	  $^1$School of Computer Science, University of Science and Technology of China, China\\ $^2$School of Biomedical Engineering, USTC, China\\ $^3$Data Darkness Lab, MIRACLE Center, Suzhou Institute for Advanced Reasearch, USTC, China\\
	  \footnotesize\texttt{\{hrwang00,yfung\}@mail.ustc.edu.cn}, 
	  \footnotesize\texttt{\{xkxie,skevinzhou\}@ustc.edu.cn}
}

\begin{document}
\maketitle
\begin{abstract}
	Although Large Language Models achieve strong success in many tasks, they still suffer from hallucinations and knowledge deficiencies in real-world applications.
	Many knowledge graph-based retrieval-augmented generation (KG-RAG) methods enhance the quality and credibility of LLMs by leveraging structure and semantic information in KGs as external knowledge bases.
	However, these methods struggle to effectively incorporate structure information, either incurring high computational costs or underutilizing available knowledge.
	Inspired by smoothing operations in graph representation learning, we propose path pooling, a simple, train-free strategy that introduces structure information through a novel path-centric pooling operation. It seamlessly integrates into existing KG-RAG methods in a plug-and-play manner, enabling richer structure information utilization.
	Extensive experiments demonstrate that incorporating the path pooling into the state-of-the-art KG-RAG method consistently improves performance across various settings while introducing negligible additional cost.
\end{abstract}

\section{Introduction}
Large Language Models (LLMs), pre-trained on vast corpora, have excelled in various natural language processing tasks \cite{shen2024understanding, naveed2023comprehensive, DBLP:conf/nips/GeHMJTXLZ23}. However, outdated information or missing domain-specific knowledge in public training corpora often lead to hallucinations in real-world applications \cite{wang2023knowledge, hong2023faithful}. To mitigate this, many methods~\citep{Pan_2024, peng2024, edge2025} integrate high-quality Knowledge Graphs (KGs) for retrieval-augmented generation (RAG), enhancing credibility.

KGs organize large collections of knowledge triples in a well-structured graph form and serve as core knowledge bases across various domains \cite{chein2008graph, robinson2015graph}.  Compared to traditional text-based knowledge bases, KGs not only capture rich semantic information but also offer a clear structure organization, enabling efficient knowledge management and updates. Therefore, KG-RAG provides higher-quality domain knowledge in many applications, significantly mitigating LLMs' hallucinations \cite{jiang-etal-2023-structgpt, knowledgeaugmented}. 

However, the complex knowledge structures in KGs pose challenges for effective knowledge retrieval in KG-RAG systems. 
Current mainstream KG-RAG methods can be categorized into \textit{path-based KG-RAG} and \textit{triple-based KG-RAG} based on their retrieval paradigms. 
Earlier path-based KG-RAG paradigm employs LLMs for path traversal on KGs \cite{ToG,ToG2.0}, subsequent path-based works try to fine-tune LLMs with KG information to retrieval knowledge graph paths \cite{RoG, GNN-RAG}. 
Although these methods effectively leverage the structure information of KGs, they face significant efficiency challenges, such as the high cost of path retrieval due to the vast number of possible path combinations in graph structures and the computational burden of multiple LLM calls during path traversal \cite{ToG,ToG2.0}, which limits their real-world applications.
To address this efficiency issue, \textit{triple-based KG-RAG} paradigm \cite{GRAG,G-Retriever,SubgraphRAG} focuses on the smallest unit in KGs—the triple. Compared to the exponential number of paths in the graph, these methods significantly improve deployment efficiency. They typically use a lightweight retriever to directly retrieve query-relevant triples, followed by a single LLM call to generate responses. Notably, the latest method, SubgraphRAG \cite{SubgraphRAG}, trains a simple multilayer perceptron (MLP) for triple retrieval based on semantic and structure information, achieving state-of-the-art performance.
Particularly in terms of efficiency, SubgraphRAG overwhelmingly outperforms previous path-based KG-RAG methods, being over 100 times faster than the representative RoG in the retrieval process while requiring only a single LLM call for generation. 

However, current triple-based KG-RAG methods, despite attempt to embed graph structure information into triple representations, primarily treat structure information as an auxiliary enhancement to semantics \cite{G-Retriever,GRAG,SubgraphRAG}. Many of these methods introduce complex structure embedding models, which significantly increase training complexity and cost. At the same time, due to the inherent nature of triples as the atomic units of a knowledge graph, they inevitably fall short in capturing structure information compared to path-based methods, where paths inherently encode richer structure context. This raises a critical question: \textbf{Can we leverage graph structure for KG-RAG in a low-cost, training-free manner within the triple-based paradigm?}

Recent studies have shown that many graph representation learning techniques \cite{hamilton2018, NEURIPS2018, 10473053} inherently perform a form of Laplacian smoothing \cite{7552590}, which aggregates features from neighboring structures. This suggests that structurally adjacent triples often share similar properties. Inspired by this structure locality pattern, we propose a path pooling strategy that leverages this smoothing effect to refine triple-based scoring, enabling richer structure information to be incorporated in a plug-and-play manner.

Our path pooling strategy addresses the limitations of current triple-based KG-RAG methods by introducing structure information through a lightweight and efficient workflow. The strategy consists of two key steps: path kernel search and smoothing along kernels. First, we identify key paths in the KGs using a graph search algorithm, such as Dijkstra's algorithm, to extract local structure patterns while avoiding over-smoothing. 
Then, we apply an average pooling operation along these paths to smooth triple scores and incorporate a position score to further preserve the structure information within the path structures.
This process enhances the representation of triples by incorporating information from their neighbors in the graph, without requiring additional training or complex embedding models. Finally, we use this smoothed triple score to optimize KG-RAG methods through reranking or reselection mechanisms, effectively improving their generation quality. This simple yet effective strategy significantly improves the accuracy of existing triple-based KG-RAG methods while introducing minimal computational overhead. By combining the efficiency of triple-based retrieval with the structure richness of path-based methods, our path pooling strategy bridges the gap between these methods, offering a novel solution for incorporating graph structure into KG-RAG systems.

\section{Preliminary}
\textbf{Knowledge Graphs (KGs).}    
KGs organize triples in a structured form, encompassing a wealth of domain-specific knowledge.
A KG can be represented by $G=\{(h, r, t) | h, t \in V, r \in E\}$, where $V$ denotes the set of entities and $E$ represents the set of relations.
Each triple $\tau=(h, r, t)$ is a fact, corresponding to an edge in the KGs, which indicates the association between entity $h$ and entity $t$ through $r$.

\noindent\textbf{KG-RAG.}    
KGs contain amount semantic information and clear structure information that can be served as high-quality external knowledge bases.
Existing research on KG-RAG mainly involves two stages: retrieval and generation.
For a given query $q$, the KG-RAG methods first retrieve knowledge $K$ from KGs relevant to $q$ in the retrieval phase: \(retrieve(q, G) \rightarrow K\).
Retrieved knowledge $K$ augments $q$, forming an enriched query $q'$:
\(prompt(q, K) \rightarrow q'\).
This $q'$ then is used as prompt \footnote{The prompt used in our experiments is detailed in the Appendix~\ref{appendix:prompt}} to enhance LLMs in the generation phase: \(Gen(q') \rightarrow answer\), improving the accuracy and credibility of the generated content.
Based on the retrieved knowledge $K$ represented in the form of paths or triples, KG-RAG can be categorized into path-based KG-RAG and triple-based KG-RAG.

\textit{Path-based KG-RAG} retrieves paths from KGs, which contain multi-hop reasoning relations between question entities and potential answer entities.
Retrieved path set can be represented as \( P = \{ p\} \), where each path is an alternating sequence of relations and entities: \(p=\langle h_1\rightarrow r_1\rightarrow h_2\rightarrow r_2\rightarrow,\dots,\rightarrow t_n\rangle\). 
Path-based KG-RAG effectively leverages the topology structure of KGs and can provide more coherent contextual information to the LLMs.

\textit{Triple-based KG-RAG} focuses on retrieving the smallest unit, triple from KGs. 
The triple-based KG-RAG methods typically assign a similarity score to each triple, thereby generating a retrieved triple sequence \(T=\{ (h_i, r_i, t_i, s_i)\}\). 
They organize the higher-scoring triples in $T$ into the form of \(\langle(h_1, r_1, t_1), (h_2, r_2, t_2), \dots, (h_n, r_n, t_n)\rangle\) to enhance the generation of LLMs.
Compared to path-based KG-RAG, triple-based KG-RAG offers higher efficiency, as well as greater flexibility and scalability. 

\section{Methodology}
\subsection{Overview}
\begin{figure*}[t]
    \centering
    \includegraphics[width=0.96\textwidth]{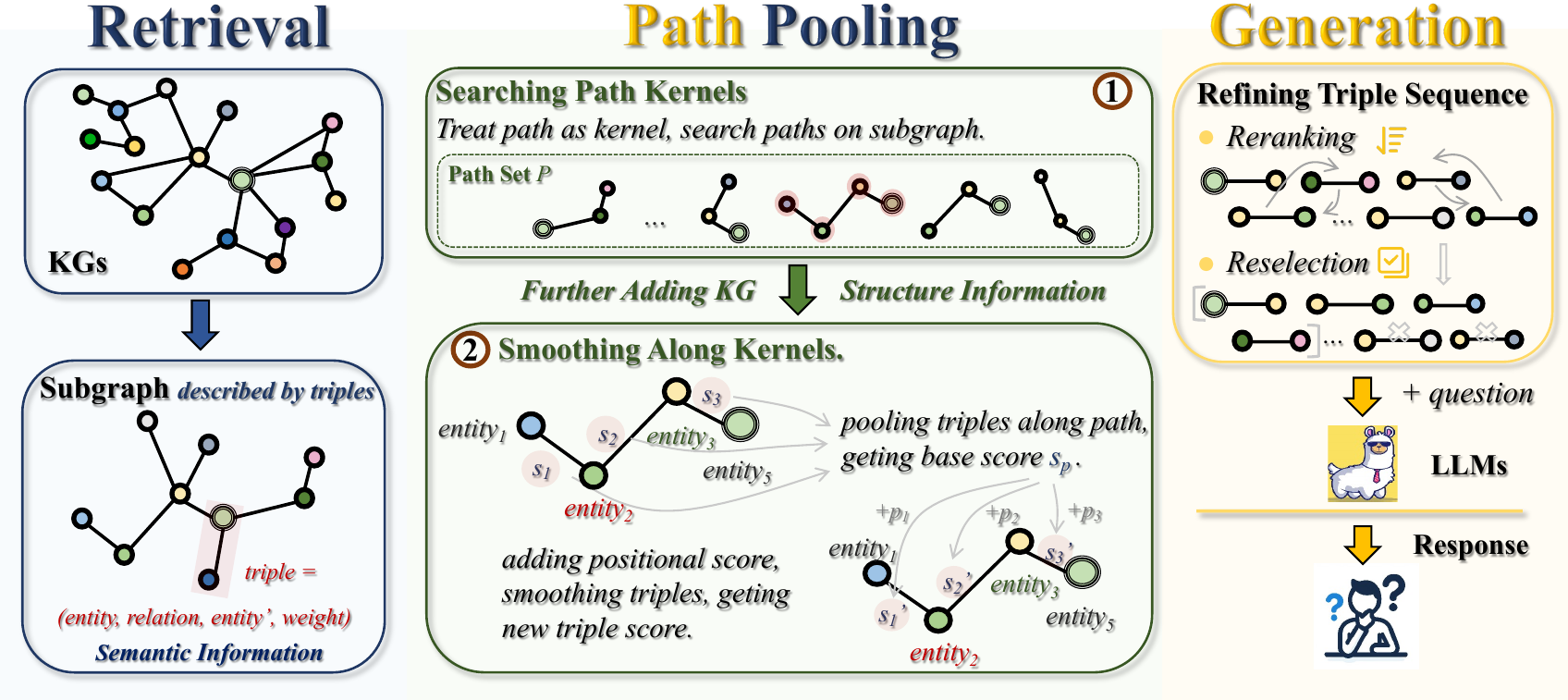}
    \caption{Framework of Path Pooling}
    \label{fig:framework}
\end{figure*}

In this section, we introduce our path pooling strategy in detail, which seamlessly bridges the retrieval and generation stages of triple-based KG-RAG. 
Path pooling contains two key steps: searching path kernels and smoothing triples along kernels.
Then, we propose reranking and reselection mechanisms to further refine the smoothed triple sequence.
The overall framework is illustrated in Figure~\ref{fig:framework}.

\subsection{Triple-based KG-RAG}
Triple-based KG-RAG methods retrieve a triple sequence $T$ from KGs by leveraging semantic and structure information of KGs jointly.
In terms of semantic information, triple-based KG-RAG methods initiate the retrieval process by encoding both the query and information of KGs into a unified vector space. 
They typically apply an off-the-shelf text encoder (e.g., SentenceBert \cite{reimers-gurevych-2019-sentence}; gte-large-en-v1.5 \cite{zhang-etal-2024-mgte}) to embed the textual attributes of entities and relations within the KGs, yielding the corresponding embeddings.
The same encoding strategy is employed to the query in order to ensure consistency.
Next, to extract the most relevant entities and relations in the KGs, triple-based KG-RAG methods employ a scorer (e.g., cosine similarity; MLP) to match semantic similarity between them and the given query.
In terms of structure information, GNNs and its variant are widely used to effective capture topology structure of KGs.  
SubgraphRAG utilizes structure information as feature to facilitate the aforementioned semantic matching process, while other methods align structure information with extracted text information to augment the LLMs.
The above retrieval process essentially functions as a pruning operation, preserving a compact triple sequence $T$ most relevant to given query for subsequently generation process.
$T$ can complement the missing knowledge of LLMs.
Algorithm~\ref{alg:triple-based_workflow} describes the workflow of triple-based KG-RAG, where path pooling effectively enhance graph structure information in a plug-and-play manner.

\subsection{Path Pooling}
\setlength{\textfloatsep}{0pt} 
\begin{algorithm}[t]
    \caption{Path Pooling Workflow}
    \label{alg:path_pooling_workflow}
    \SetAlgoLined
    \KwData{Triple sequence \(T=\{\tau=(h,r,t,s)\}\)}
    Construct subgraph \( G_k = (V_k, E_k) \) from \( T \)\;
    Search path kernels on \( G_k \)\;
    \(P=\{p_{\{e\} \rightarrow v} | v \in V_k\} \cup \{p_{v \rightarrow \{e\}} | v \in V_k\}\)\;
    \ForEach{path \( p=(\tau_1, \tau_2, \ldots, \tau_l) \in P \)}{
        Calculate path kernel score by pooling\;
        \(s_p = \text{pooling}(p)=\frac{1}{l}\sum_{i=1}^{l}s_i\)\;
    }
    
    Assign positional score \(p_i\) to \(\tau_i\)\;
    \(p_i=\frac{\min(\{s_j | \tau_j \in T\})}{i\cdot a}, S(\tau_i) = s_p + p_i\)\;
    \(s_i' = \max\{S(\tau_i) | \tau_i \in p, p\in P\}\)\;
    \Return Smoothed Triple sequence \(T'\)\;
\end{algorithm}
\setlength{\textfloatsep}{5pt}

\setlength{\textfloatsep}{0pt} 
\begin{algorithm}[t]
    \caption{Triple-based KG-RAG}
    \label{alg:triple-based_workflow}
    \SetAlgoLined
    \KwData{KG \(G=\{(h,r,t)\}\), given query \(q\)}
    Get subgraph from \(G\) centered on query entities\;
    Retrieve triple sequence $T$ on the subgraph\;
    \If{enhanced with path pooling}{
        $T'\leftarrow$ Algorithm~\ref{alg:path_pooling_workflow}($T$)\;
        Rerank or Reselect the refined $T'$ to $T''$\;
    }
    \(q' \leftarrow prompt(q, T'')\), \(answer \leftarrow Gen(q')\)\;
    \Return $answer$\;
\end{algorithm}
\setlength{\textfloatsep}{5pt}

Existing KG-RAG methods struggle to fully utilize knowledge graph (KG) structures. Path-based KG-RAG paradigm introduces structure through traversal and scoring but incur high computational costs, while triple-based KG-RAG paradigm encodes structure information yet primarily serve as semantic enhancements, leaving the graph's structure underutilized. Inspired by the common smoothing operation in graph representation learning, we propose a path pooling strategy that refines triple-based scoring by leveraging this smoothing effect. We first identifies key paths using a search algorithm, then smooths scores along these paths to enhance structure information. Surprisingly, this simple yet effective strategy significantly improves the accuracy of existing triple-based KG-RAG methods while introducing minimal additional overhead.
Algorithm~\ref{alg:path_pooling_workflow} describes the details of path pooling's workflow.

\textbf{Searching Path Kernels.}
To capture the inherent locality of graph structures, we apply graph search algorithms (e.g., Dijkstra; BFS; Random Walk) on $G_k$ constructed from triple sequence $T$, to extract paths composed of triples. Given the common issue of over-smoothing in graph learning \cite{Chen_Lin_Li_Li_Zhou_Sun_2020, Bo_Wang_Shi_Shen_2021, rusch2023}, we employ Dijkstra's algorithm to find the shortest paths starting from the query entities. 
Compared to more complex path search algorithms, shortest path search algorithm is more efficient and keeps computational costs manageable \footnote{In experiments Section~\ref{sec:path_search}, we also explored other graph algorithms for path kernel search, all of which led to performance improvements within our path pooling strategy.}. 
For triples unreachable from or to query entities, we treat them as single-hop paths for subsequent unified processing.   
So far, we obtain the path kernel set $P$.

\textbf{Smoothing Along Kernels.}
Next, we apply a smoothing operation to the triples along the searched path kernel. In traditional graph representation learning, this process often relies on learnable weighted aggregation, which provides finer-grained information smoothing but typically incurs high training costs. Given the efficiency concerns in KG-RAG, we adopt mean smoothing, similar to traditional average pooling. Specifically, we take the average score along the path kernel as the base score for all triples.
Average score integrally considers all triples on a path, providing excellent smoothing effect, demonstrating good numerical stability, and exhibiting strong resistance to the influence of outliers.
The smoothing operation compensates the KGs' structure information of the retrieved triples to a certain extent, allowing triples in the same path to have more similar scores.

Additionally, we introduce a fine-grained positional score to better preserve positional information along the path. 
For $T$, we first calculate the minimum $s_{min}$ of all triples.
Based on the order triples appear in the path, we sequentially assign additional score to them, the first triple adding $s_{min}$ divided by $a$, the second $s_{min}$ divided by $2a$, and so forth, where a is a nonzero constant.
In practice, different paths in the path set $P$ may contain same triples.
To address this issue, we adopt maximum as the final triple score. 
Through this process, a triple's representation is smoothed via its neighbors' representations, since each triple is able to receive additional information from neighboring triples, while maintaining the relative position order within the path.  

\subsection{Boosting KG-RAG with Path Pooling}

After path pooling which effectively incorporates structure information, we obtain a smoothed triple sequence $T'=\{\tau_i' | \tau_i'=(h_i, r_i, t_i, s_i')\}$.
We then demonstrate two ways to further refine the triple sequence $T'$ to enhance triple-based KG-RAG: Position Reranking and Triple Reselection.

\textbf{Position Reranking.} Due to the position encoding, LLMs exhibit significant sensitivity to token positions, a phenomenon known as position bias \cite{tang2023found, hsieh2024found, zhang2024attention}. 
This bias typically manifests in two cases: 1) \textit{Lost in the Middle}~\cite{lostInMiddle, xu2024retrieval, cao2024, jin2025longcontext}: This typically occurs in models with limited long-sequence capabilities, where tokens in the middle of the input are often overlooked during inference. 
2) \textit{Recency Bias}: \cite{qin-etal-2023-nlp, peysakhovich2023}
This is prevalent in most models and is primarily attributed to the inherent properties of popular rotary position encoding. 
Tokens appearing later in the sequence tend to receive higher attention weights. Many RAG studies \cite{peysakhovich2023, he2024} have shown that reranking retrieved content based on position bias can effectively improve the quality of model-generated responses. Following this insight, we leverage the smoothed triple scores from path pooling to reorder retrieved triples, ensuring a better optimal input sequence $T''$. Specifically, we position triples with higher scores closer to the final question, adapting the widely observed \textit{Recent Bias}. This mechanism enhances existing KG-RAG methods by optimizing the input order of triples without altering their content, solely relying on smoothed scores for reordering.

\textbf{Triple Reselection.} 
Another key challenge in KG-RAG is balancing high-quality retrieval context while minimizing the inclusion of question-irrelevant tokens. Removing such irrelevant tokens not only reduces noise interference but also lowers inference costs. Inspired by the widely adopted coarse-to-fine selection paradigm \cite{yu2024rankrag, zhang2024cream, zhao2025funnel}, we propose Triple Reselection to enhance performance. Specifically, we first apply the naive triple-based KG-RAG scorer for a coarse-grained selection, retaining the top-$k$ high-scoring triples. 
Then, we refine this selection by applying path pooling to obtain more precise scores, further filtering down to the top-$k'$ triples.  Finally, these fine-grained selected triple sequence $T''$ is fed into the LLM in an order aligned with the \textit{Recent Bias}. 
This mechanism not only optimizes position ranking using smoothed scores but also filters out irrelevant information, leading to better performance.

\section{Experiments}
\subsection{Experiment Setup}
\textbf{Dataset.}
To evaluate the effectiveness of our path pooling strategy in multi-hop knowledge-intensive tasks, all of our experiments are based on the widely-used CWQ dataset, which is constructed from the Freebase \cite{Freebase}. 
Freebase is an rich knowledge graph, containing 126 million triples.
CWQ \cite{CWQ} is a challenging benchmark for KGQA that includes reasoning questions involving up to four hops.

\noindent\textbf{Evaluation Metrics.}
Consistent with previous works \cite{RoG,GNN-RAG,SubgraphRAG}, we adopt Macro-F1 and Hit@1 as our evaluation metrics. Hit@1 measures whether the correct answer appears as the top-ranked result. Macro-F1, which accounts for both precision and recall in scenarios with multiple valid answers, provides a more comprehensive assessment of reasoning performance

\noindent\textbf{Implementation Details.} 
We enhance the state-of-the-art triple-based KG-RAG method, SubgraphRAG, using path pooling through positional reranking and triple reselection.
For brevity, we abbreviate SubgraphRAG as S-RAG in the following experiments.
To evaluate the performance of our path pooling strategy, we employ a diverse selection of open-source LLMs: Llama3.1-8b-Instruct, Llama3.1-70b-Instruct \cite{llama3.1}, Qwen2.5-7b-Instruct, and Qwen2.5-72b-Instruct \cite{qwen2025}. For all models, we use a one-shot prompt for reasoning, as detailed in Appendix~\ref{appendix:prompt}.
For all LLMs, we employed one-shot prompt for reasoning. 
The reasoning prompt is detailed in Appendix~\ref{appendix:prompt}.
To ensure reproducibility, we set the temperature to 0 and the maximum token length for generation to 4000, aligning with previous research \cite{SubgraphRAG}.
\subsection{Main Results}
\begin{table}[t]
    \small
    \centering
    \setlength{\tabcolsep}{2pt} 
    \begin{tabular}{l|c|c|c|c|c|c}
        \toprule
        \multicolumn{7}{c}{\textbf{Triple Num 25\hspace{1em}+1.6ms/query({\scriptsize{PR}}) \hspace{1em}+18.4ms/query({\scriptsize TR})}} \\
        \midrule
        Methods & \multicolumn{3}{c|}{F1-Score} & \multicolumn{3}{c}{Hit@1} \\
        \cline{2-7}
                    & w/o ours   & w/ \textbf{{\scriptsize PR}}  & w/ \textbf{{\scriptsize TR}}    & w/o ours  & w/ \textbf{{\scriptsize PR}}      & w/ \textbf{{\scriptsize TR}}\\
        \midrule
        Llama-8b    & 45.4  & \underline{46.0} & \textbf{48.4} & 48.7 & \underline{49.2} & \textbf{52.4}\\
        Llama-70b   & 52.3  & \underline{54.0} & \textbf{55.8} & 56.0 & \underline{58.7} & \textbf{61.2}\\
        Qwen-7b     & 40.7  & \underline{42.2} & \textbf{45.0} & 43.8 & \underline{46.2} & \textbf{49.1}\\
        Qwen-72b    & 49.3  & \underline{51.3} & \textbf{53.6} & 52.5 & \underline{54.9} & \textbf{57.0}\\
        \midrule
        \multicolumn{7}{c}{\textbf{Triple Num 50\hspace{1em}+2.1ms/query({\scriptsize{PR}}) \hspace{1em}+18.4ms/query({\scriptsize TR})}} \\
        \midrule
        Methods & \multicolumn{3}{c|}{F1-Score} & \multicolumn{3}{c}{Hit@1} \\
        \cline{2-7}
                    & w/o ours   & w/ \textbf{{\scriptsize{PR}}}  & w/ \textbf{{\scriptsize TR}}    & w/o ours  & w/ \textbf{{\scriptsize PR}}      & w/ \textbf{{\scriptsize TR}}\\
        \midrule
        Llama-8b    & 46.8 & \underline{47.8} & \textbf{50.5} & 49.6 & \underline{51.7} & \textbf{54.0}\\
        Llama-70b   & 53.6 & \underline{54.5} & \textbf{56.5} & 57.5 & \underline{58.9} & \textbf{60.9}\\
        Qwen-7b     & 41.9 & \underline{43.3} & \textbf{46.0} & 45.2 & \underline{47.1} & \textbf{50.1}\\
        Qwen-72b    & 51.2 & \underline{53.1} & \textbf{55.6} & 54.3 & \underline{56.4} & \textbf{59.0}\\
        \midrule
        \multicolumn{7}{c}{\textbf{Triple Num 100\hspace{1em}+3.2ms/query({\scriptsize PR}) \hspace{1em}+18.5ms/query({\scriptsize TR})}} \\
        \midrule
        Methods & \multicolumn{3}{c|}{F1-Score} & \multicolumn{3}{c}{Hit@1} \\
        \cline{2-7}
                    & w/o ours   & w/ \textbf{{\scriptsize PR}}  & w/ \textbf{{\scriptsize TR}}    & w/o ours  & w/ \textbf{{\scriptsize PR}}      & w/ \textbf{{\scriptsize TR}}\\
        \midrule
        Llama-8b    & 47.7 & \underline{48.8} & \textbf{51.0} & 50.2 & \underline{52.7} & \textbf{54.4}\\
        Llama-70b   & 53.0 & \underline{54.2} & \textbf{56.5} & 57.2 & \underline{58.9} & \textbf{61.2}\\
        Qwen-7b     & 42.8 & \underline{44.0} & \textbf{46.1} & 45.8 & \underline{47.4} & \textbf{49.9}\\
        Qwen-72b    & 52.1 & \underline{53.5} & \textbf{55.4} & 55.3 & \underline{57.0} & \textbf{59.1}\\
        \midrule
        \multicolumn{7}{c}{\textbf{Triple Num 200\hspace{1em}+6.3ms/query({\scriptsize PR}) \hspace{1em}+18.5ms/query({\scriptsize TR})}} \\
        \midrule
        Methods & \multicolumn{3}{c|}{F1-Score} & \multicolumn{3}{c}{Hit@1} \\
        \cline{2-7}
                    & w/o ours   & w/ \textbf{{\scriptsize PR}}  & w/ \textbf{{\scriptsize TR}}    & w/o ours  & w/ \textbf{{\scriptsize PR}}      & w/ \textbf{{\scriptsize TR}}\\
        \midrule
        Llama-8b    & 47.2 & \underline{49.5} & \textbf{50.2} & 50.1 & \underline{53.2} & \textbf{54.2}\\
        Llama-70b   & 52.8 & \underline{54.5} & \textbf{55.5} & 57.6 & \underline{59.6} & \textbf{60.7}\\
        Qwen-7b     & 41.8 & \underline{43.4} & \textbf{45.1} & 44.9 & \underline{46.9} & \textbf{48.7}\\
        Qwen-72b    & 51.9 & \underline{53.7} & \textbf{54.6} & 55.6 & \underline{57.3} & \textbf{58.7}\\
        \bottomrule
    \end{tabular}
    \caption{Results of Positional Reranking and Triple Reselection with Path Pooling. The best scores are highlighted with \textbf{bold} and the second-best scores are highlighted with \underline{underline}. \textbf{PR} denotes our positional reranking mechanism with path pooling and \textbf{TR} denotes our triple reselection mechanism with path pooling.}
    \label{tab:rank_results}
\end{table}
\begin{figure}[ht]
    \centering
    \includegraphics[width=0.5\textwidth]{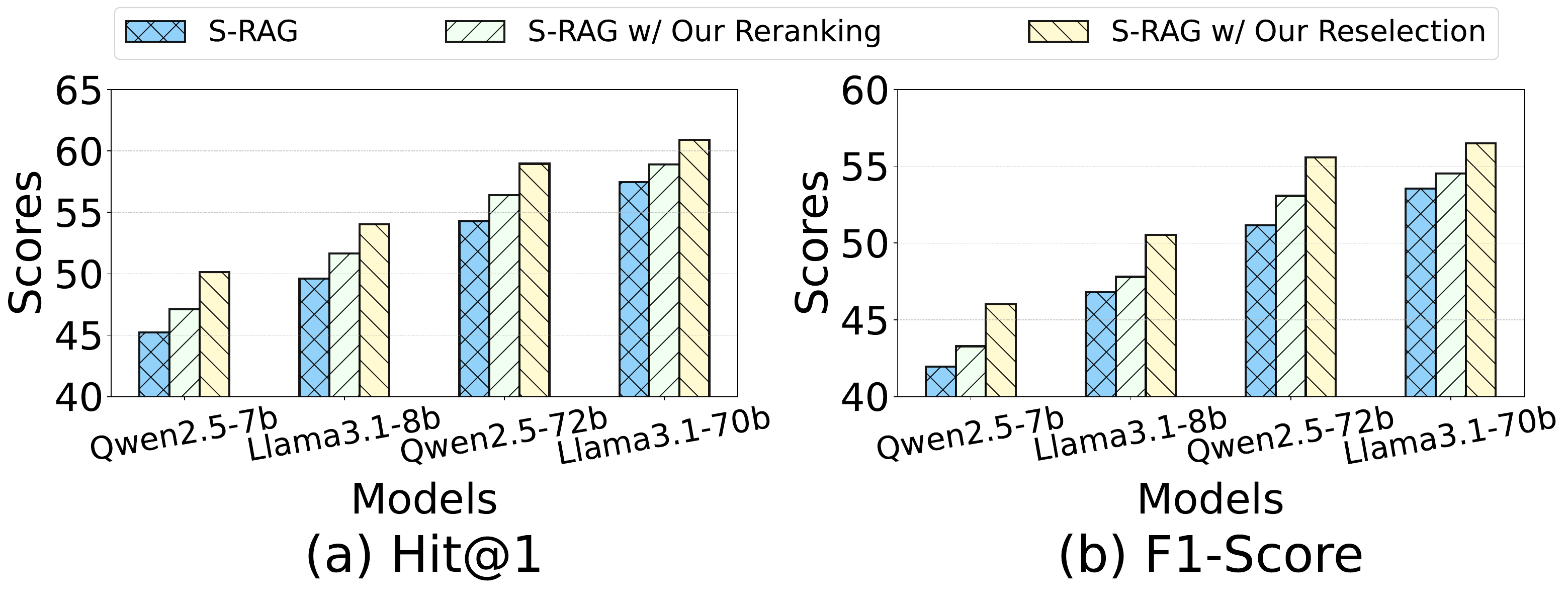} 
    \caption{Results of Different Models with 50 Triples}
    \label{fig:main_performance}
\end{figure} 

The experimental results of the reranking mechanism, which refines the triple sequence using path pooling, are presented in Table~\ref{tab:rank_results}. Overall, reranking introduces minimal additional retrieval time, as it only requires reordering triples based on the refined scores from path pooling. For instance, even when processing up to 200 triples, the additional retrieval time is only 3.2 ms per query, while achieving a 3.1\% improvement in Hit@1 and a 2.3\% increase in F1-Score on Qwen2.5-7b. 
Furthermore, when applying path pooling for triple reselection, the performance gains become even more pronounced. We conduct experiments by first retrieving 500 triples using SubgraphRAG and then selecting 25 to 200 triples based on their path pooling scores. Notably, with just 25 triples, the reselection mechanism boosts Hit@1 by 5.2\% on Llama3.1-70b and 5.3\% on Qwen2.5-7b while introducing only 18.4ms per query.
Remarkably, using only 25 triples, the reselection mechanism even outperforms the original SubgraphRAG utilizing 200 triples, demonstrating its high effectiveness.

\begin{figure}[ht]
    \centering
    \includegraphics[width=0.5\textwidth]{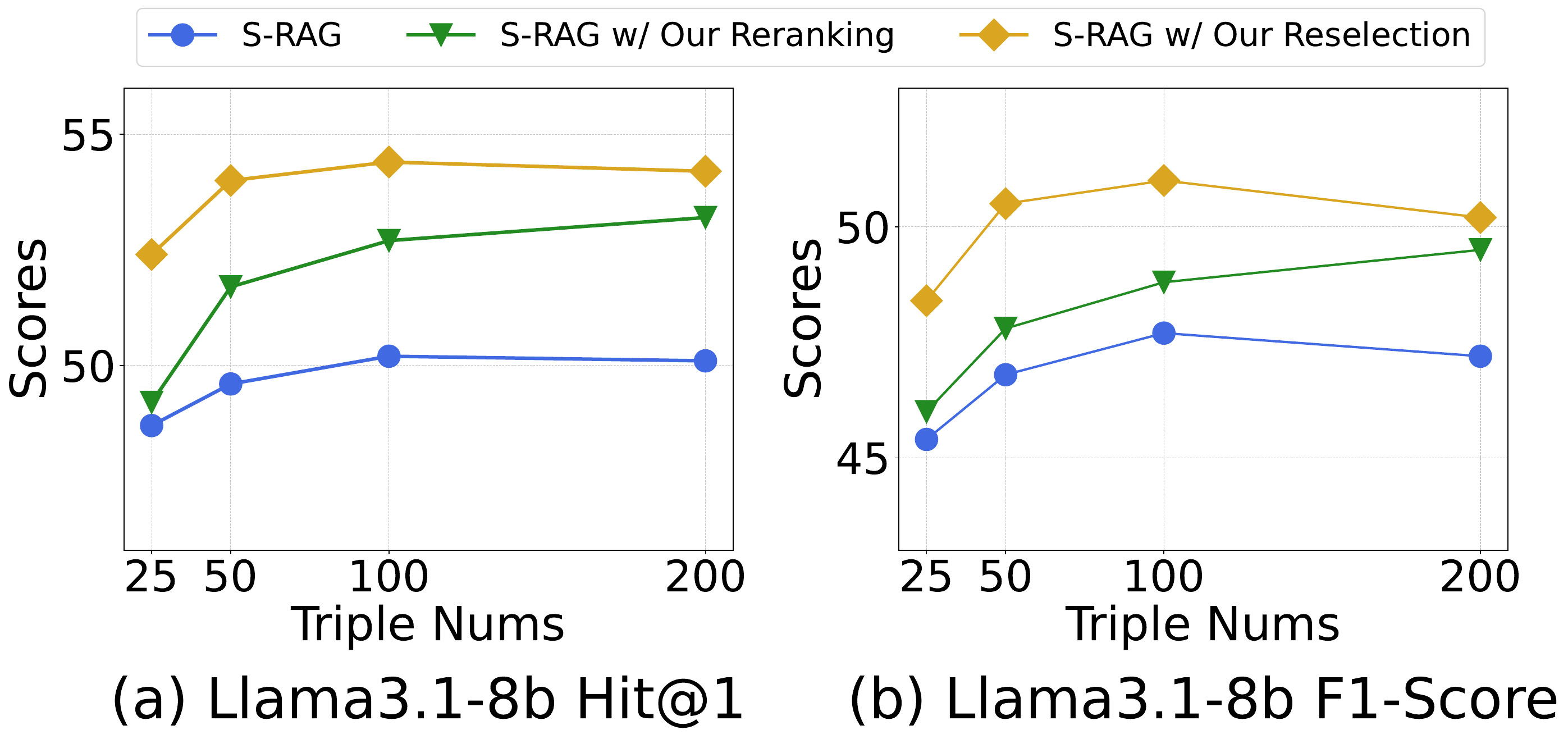} 
    \caption{Results of Varying Triples}
    \label{fig:line_performance}
\end{figure}

Figure~\ref{fig:main_performance} provides a detailed visualization of this phenomenon in triple num 50. Our path pooling strategy consistently enhances the performance of SubgraphRAG across various settings. Simply leveraging the structure information from path pooling to rerank triples in the prompt based on positional bias already leads to notable improvements. Moreover, refining triple reselection using path pooling scores to filter out irrelevant noise results in further performance gains.
Figure~\ref{fig:line_performance} further visualizes the performance of Llama 3.1-8B \footnote{In the paper of SubgraphRAG, the Hit@1 of Llama3.1-8b-Instruct with 100 triples is 56.98.} as the number of triples increases from 25 to 200. When using path pooling for reranking, performance consistently improves with more triples, achieving substantial gains at 200 triples, with Hit@1 and Macro-F1 increasing by. This aligns with prior findings on positional bias: the longer the sequence, the more pronounced the bias \cite{lostInMiddle}. In contrast, when applying triple reselection, path pooling achieves significant improvements with as few as 25 triples, boosting Hit@1 and Macro-F1 by. However, as the number of triples increases further, the performance gains gradually diminish.

\subsection{Further Analysis}
We conduct additional experiments to evaluate the impact of different path search and pooling algorithms in our path pooling strategy.
For demonstration, we apply the triple reselection mechanism with path pooling on Llama3.1-8b.
And the results of positional reranking mechanism with path pooling are reported in Appendix~\ref{appendix:analysis}.

\subsubsection{Different Path Search Algorithms}
\label{sec:path_search}

\begin{figure}[ht]
    \centering
    \includegraphics[width=0.5\textwidth]{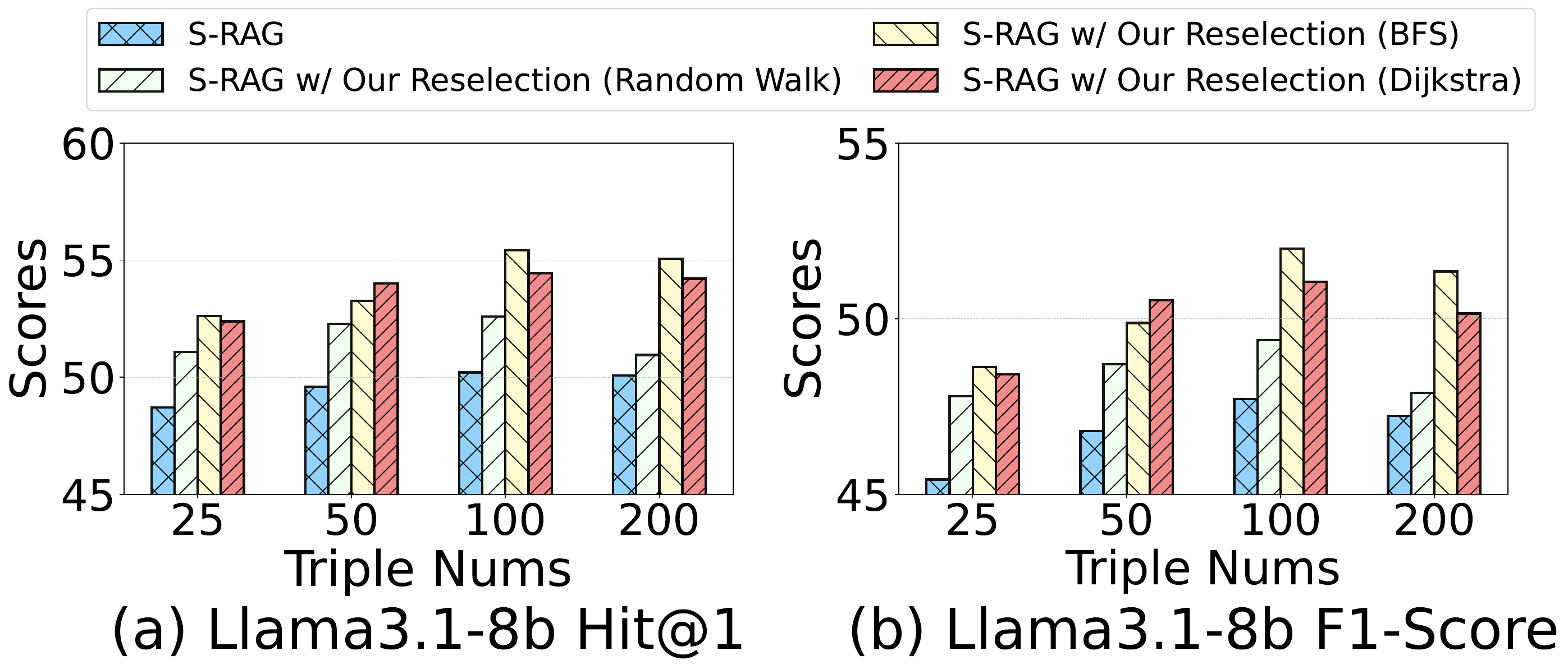}
    \caption{Reselection Results of Different Path Search Algorithms on Llama3.1-8b}
    \label{fig:path_search}
\end{figure}

Beyond the widely used shortest path algorithm, Dijkstra, other graph traversal methods such as BFS and Random Walk can also be employed to extract structure information.  
We evaluate the impact of different path search algorithms within our path pooling strategy on  Llama3.1-8b, as shown in Figure~\ref{fig:path_search}.  
To prevent excessive smoothing of structural information, we set the maximum path length for BFS and Random Walk to 4, and Random Walk samples 256 times.  
Overall, all path search algorithms contribute to performance improvements when integrated into path pooling. However, Random Walk performs significantly worse than the other two, likely because its stochastic nature fails to capture key structure information.  
While BFS and Dijkstra achieve similar performance, BFS introduces a significantly larger number of paths, leading to increased computational overhead, as shown in Table~\ref{tab:time_statistics}.  
For instance, with 500 triples, BFS requires 232.2 ms, whereas Dijkstra only takes 18.1 ms, making it 12.8x times more efficient.  
Thus, we recommend using Dijkstra in the path pooling strategy to achieve a balance between effectiveness and efficiency.

\begin{table}[t]
    \centering
    \small
    \begin{tabular}{@{}lrrrrr@{}}
    \toprule
    Triple Num      & 25    &50     &100    &200    &500   \\\midrule
    Dijkstra        & 1.6   &2.1    &3.2    &6.3    &18.1   \\
    Random Walk     & 3.1   &3.6    &4.5    &5.7    &9.1   \\
    BFS             & 1.9   &3.1    &7.4    &29.4   &232.2  \\\bottomrule
    \end{tabular}
    \caption{Time Statistics of Different Path Search Algorithms (ms/query)}
    \label{tab:time_statistics}
\end{table}

\subsubsection{Different Pooling Strategies}
\begin{figure}[ht]
    \centering
    \includegraphics[width=0.5\textwidth]{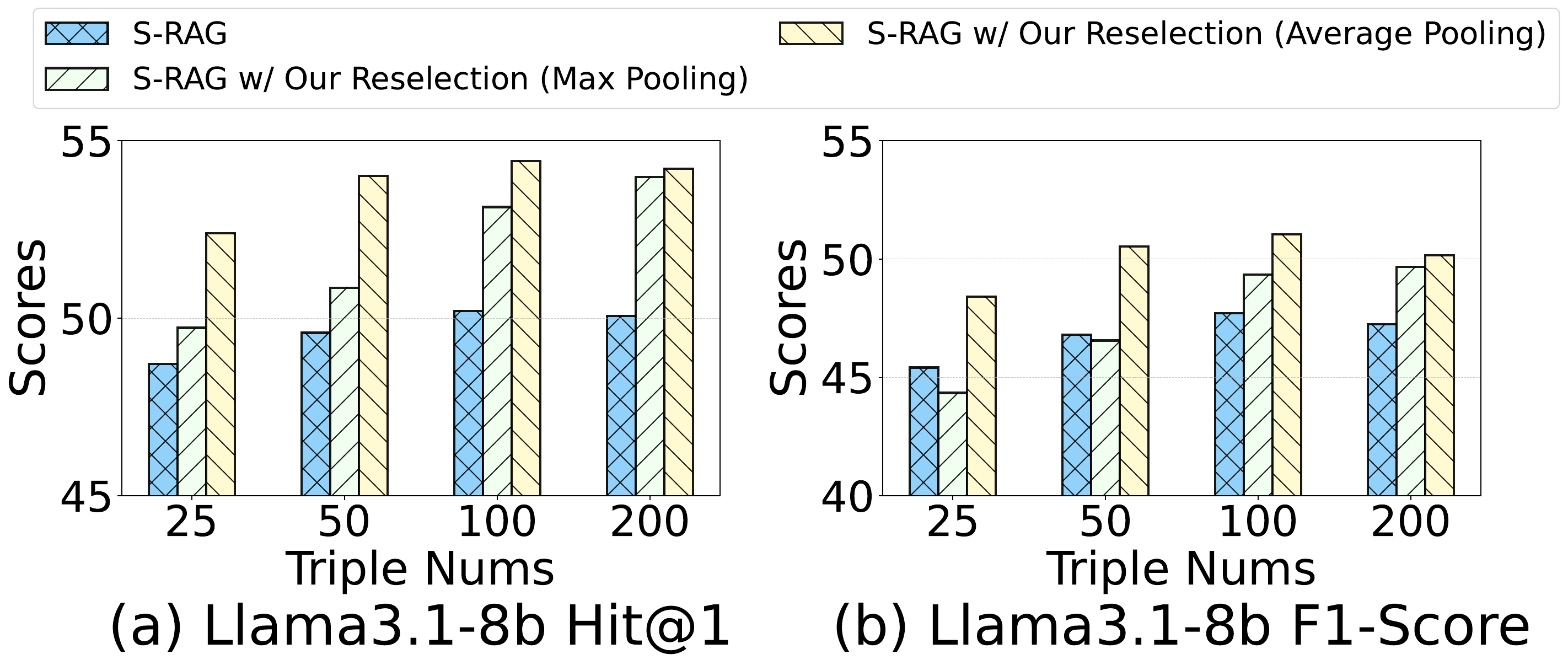}
    \caption{Reselection Results of Different Pooling Strategies on Llama3.1-8b}
    \label{fig:pooling_strategy}
\end{figure}

In Figure~\ref{fig:pooling_strategy}, we further compare the performance of different smoothing operations in our path pooling strategy. We evaluate two commonly used pooling operation for smoothing: max pooling, which retains the maximum score along the path, and average pooling, which computes the mean score of triples on the path as the base score.
Both operation could capture local graph structure information, enhancing the reasoning performance of LLMs. However, average pooling consistently outperforms max pooling, especially when the number of triples is small. For instance, with 50 triples, average pooling shows a significant advantage over max pooling. Using max pooling provides a increase of 1.27\%, from 0.5086 to 0.4959, while leveraging average pooling brings a substantial improvement of 4.42\%, from 0.5401 to 0.4959. This suggests that average pooling better preserves neighborhood information, making it the preferred choice for smoothing in path pooling.

\subsubsection{Lost in the Middle Positional Bias}
\begin{figure}[ht]
    \centering
    \includegraphics[width=0.5\textwidth]{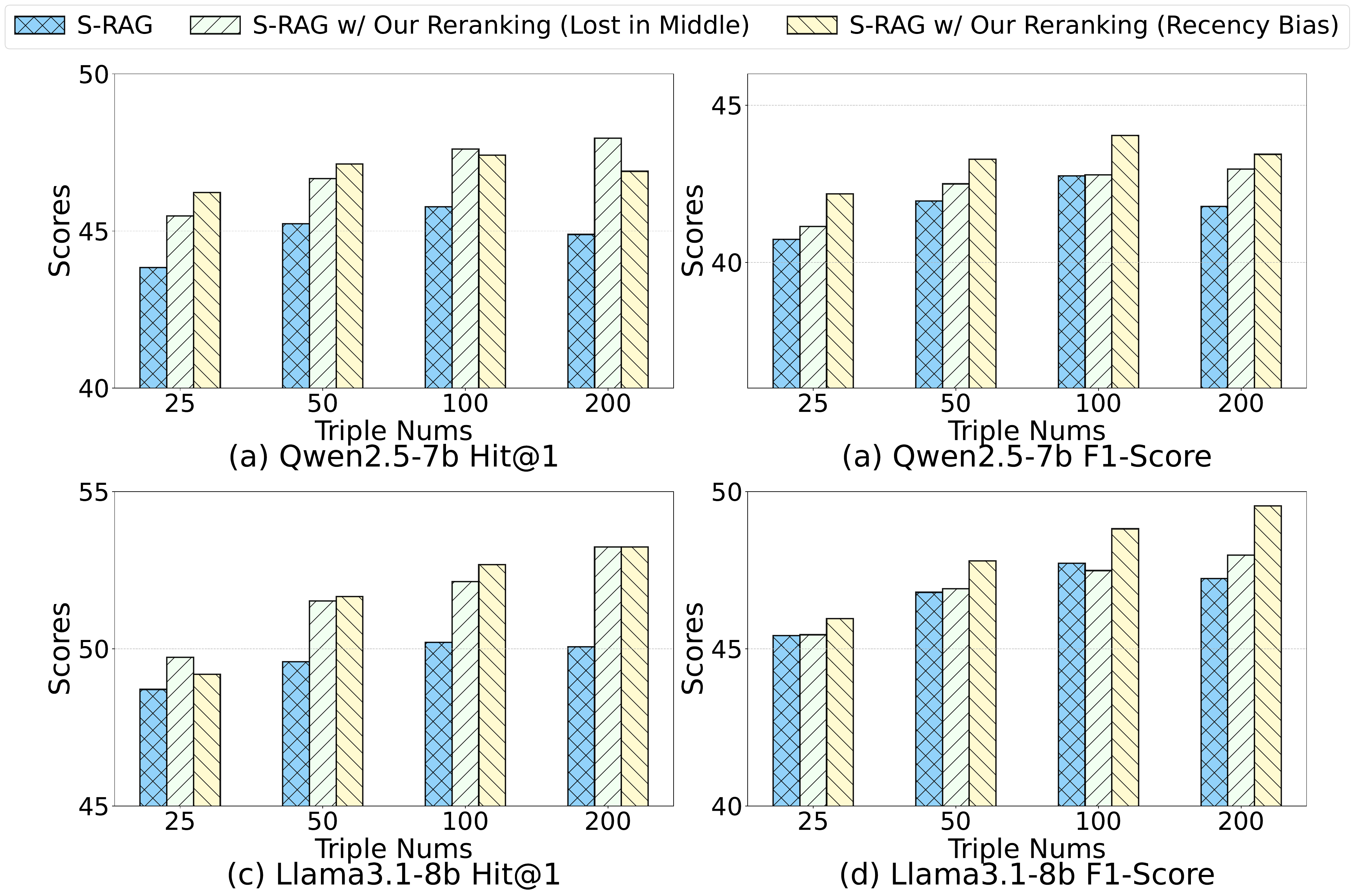}
    \caption{Reranking Results of Lost in Middle on Llama3.1-8b}   
    \label{fig:lost_in_middle_reranking}
\end{figure}
We further explore the effect of reranking triples based on lost in the middle positional bias, as shown in Figure~\ref{fig:lost_in_middle_reranking}.
Starting from the triples with the highest smoothed scores, we iteratively place them in the head and tail of the input sequence.
Overall, both Lost in the Middle and Recency Bias demonstrate significant improvements over SubgraphRAG. Specifically, Lost in the Middle shows slightly weaker performance on Qwen, while achieving comparable results on Llama. This discrepancy is primarily due to differences in positional bias across models. Future work could further optimize performance by investigating these positional biases in greater depth. Given its simplicity and robustness, we recommend incorporating Recency Bias into path pooling.

\subsubsection{Positional Reranking with More Triples}
\begin{table}[t]
    \small
    \centering
    \setlength{\tabcolsep}{2pt} 
    \begin{tabular}{l|cc|c|cc|c}
        \toprule
        \multicolumn{7}{c}{\textbf{Triple Num 500\hspace{1em}+18.1ms/query}} \\
        \midrule
        Methods & \multicolumn{3}{c|}{F1-Score} & \multicolumn{3}{c}{Hit@1} \\
        \cline{2-7}
        & w/o ours & w/ \textbf{PR} & \(\Delta\) & w/o ours & w/ \textbf{PR} & \(\Delta\) \\
        \midrule
        Llama-8b & 45.5 & \textbf{48.3} & +2.8 & 49.2 & \textbf{53.3} & +4.1  \\
        Llama-70b & 51.8 & \textbf{53.6} & +1.8 & 56.9 & \textbf{58.7} & +1.8  \\
        Qwen-7b & 41.7 & \textbf{43.5} & +1.8 & 44.9 & \textbf{47.1} & +2.2  \\
        Qwen-72b & 52.2 & \textbf{54.7} & +2.5 & 55.8 & \textbf{58.8} & +3.0  \\
        \bottomrule
    \end{tabular}
    \caption{Reranking Results with 500 triples}
    \label{tab:rank_results}
\end{table}
Table~\ref{tab:rank_results} presents the results of the positional reranking mechanism with path pooling using more triples.
As the number of triples expands, for instance to 500, the corresponding increase in token counts introduces challenges associated with long sequence reasoning. 
Despite this, SubgraphRAG, when augmented with our positional reranking mechanism, demonstrates notable performance advantages. 
This is particularly evident in the results for Llama3.1-8b. 
Specifically, leveraging our positional reranking mechanism yields a 4.1\% improvement in Hit@1 and a 2.8\% increase in F1-Score.

\section{Related Work}
\noindent\textbf{Path-based KG-RAG }    
retrieves top-$k$ knowledge graph paths relevant to the query, fully utilizing the structure information of KGs.
Existing research has focused on retrieves paths to provide updated or domain-specific contextual knowledge for LLMs.
For instance, GNN-RAG \cite{GNN-RAG} trains a GNN to retrieve answer candidates for a given query from a dense KG subgraph and extracts the shortest paths between the question entities and the answer candidates. 
RoG \cite{RoG} utilizes a planning-retrieval-reasoning framework to fine-tune the LLMs for generating relation paths, and then retrieves complete paths from the KG for LLMs reasoning.
Nevertheless, GNN-RAG and RoG both face signiﬁcant efficiency challenges, as they fine-tune LLMs, resulting in high training overhead and retrieval latency.
Subsequent, agent-based method, ToG \cite{ToG} attempt to leverage LLMs to dynamically retrieve paths from KGs and make decisions accordingly.
ToG 2.0 \cite{ToG2.0} proposes a hybrid RAG paradigm that effectively combines unstructured text information with structured graph information, enabling depth and comprehensive retrieval processes.
However, ToG and ToG 2.0 tightly couple LLMs with retrieval process, requiring multiple calls, which leads to high computitional burden.

\noindent\textbf{Triple-based KG-RAG }
retrieves the top-$k$ triples relevant to the query, which is more efficient compared to path-based KG-RAG. GRAG \cite{GRAG} employs a divide-and-conquer strategy to retrieve subgraphs that are most relevant to the query, utilizing a graph pruning mechanism to unify and refine them. During the reasoning phase, GRAG provides the LLMs with both text and graph views of information, enabling a deeper understanding of the relations between entities.
G-retrieval \cite{G-Retriever} integrates the advantages of LLM, GNN and RAG, which employs cosine similarity to retrieve entities and relations, while the PCST algorithm manages the size of the subgraph. The textual representations of the subgraph, along with the structure information outputted by the GNN, are fed into the LLM to generate answers.
Unlike previous work, currently SubgraphRAG \cite{SubgraphRAG} trains a lightweight multilayer perceptron, which enables scalable and flexible retrieval processes, achieving higher retrieval efficiency and better performance.
Compared to path-based KG-RAG, none of the above methods require fine-tuning or multiple calls to the LLMs. 
Our path pooling strategy can be seamlessly pluged into the retrieval and generation phases of the triple-based KG-RAG in a low-cost, training-free manner.
It enhances the performance of LLMs by introducing graph structure information in the retrieved triple sequence.

\section{Conclusion}

In this paper, we address a key challenge in KG-RAG: integrating rich structure information from knowledge graphs without incurring excessive computational costs. Current triple-based KG-RAG methods are efficient but fail to capture the full graph structure, while path-based KG-RAG methods, though more expressive, are computationally expensive. To bridge this gap, we propose the \text{path pooling} strategy to enhance existing triple-based KG-RAG in a plug-and-play manner by leveraging graph smoothing effects.
Our method identifies key paths using graph search (e.g., Dijkstra’s algorithm) and applies smoothing to refine triple scores along these paths, effectively enriching the structure representation in triple-based KG-RAG. 
Experiments demonstrate that path pooling consistently improves KG-RAG performance with minimal overhead. 
This work introduces a novel \textit{training-free} enhancement strategy for KG-RAG by leveraging smoothing effects widely used in graph learning.
Future work could explore more effective path searching and finer-grained smoothing techniques to capture richer structure information.

\section*{Limitations}
While path pooling effectively leverages paths searched from query entities to smooth retrieved triples, thereby addressing the underutilization of structural information in KG-RAG, its full potential remains underexplored. 
Our current experiments predominantly relied on conventional path search algorithms, such as BFS or Dijsktra, without investigating potentially more efficacious weighted or heuristic path searching methods. 
Furthermore, introducing structural information via paths for triple smoothing represents only one specific method.
KGs possess richer structural characteristics, such as node degree, adjacency relationships, and local subgraph topologies, which offer promising yet unexplored avenues for effective triple smoothing and warrant future investigation.

\bibliography{custom}
\appendix
\section{Appendix}
\label{appendix:analysis}
\subsection{Triple Reselection on Qwen2.5-7b}
\subsubsection{Different Path Search Algorithms}
\begin{figure}[ht]
    \centering
    \includegraphics[width=0.5\textwidth]{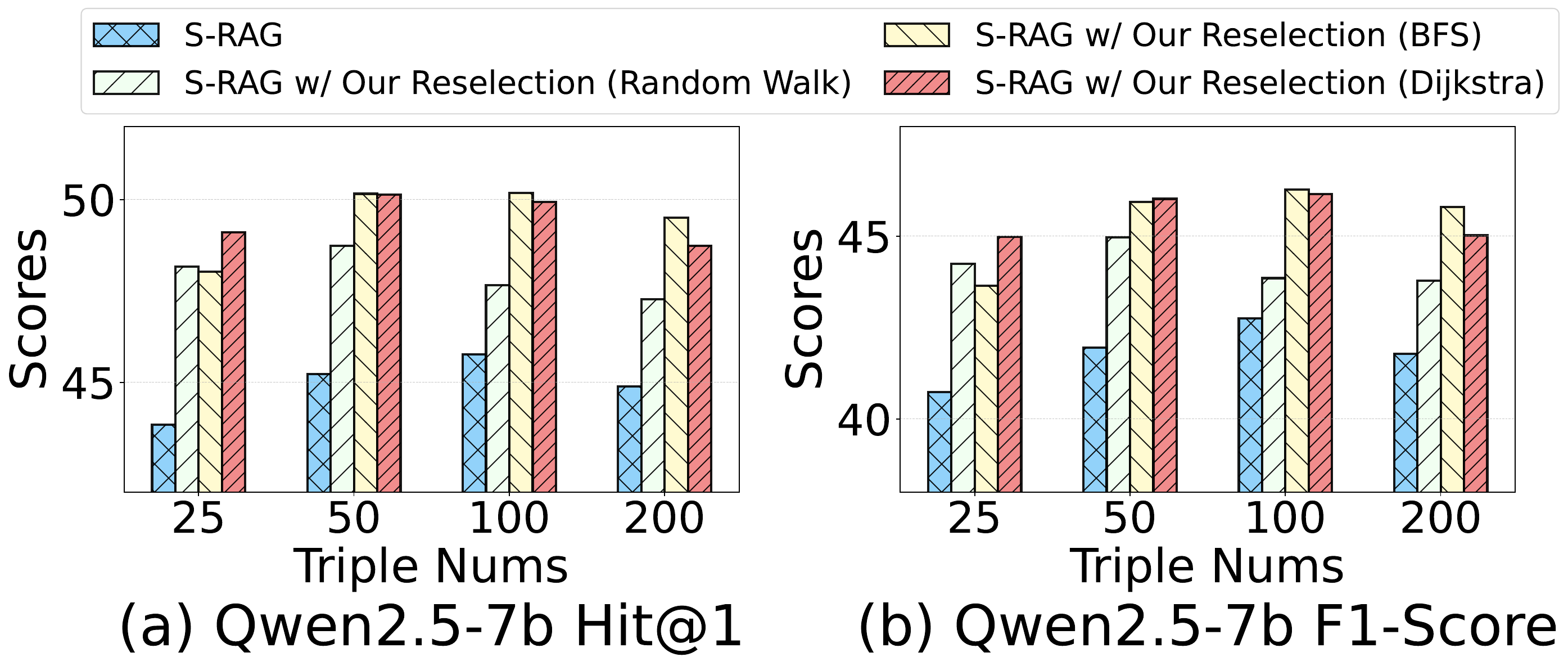}
    \caption{Reselection Results of Path Search Algorithms on Qwen2.5-7b}
    \label{fig:qwen_path_search}
\end{figure}

Triple reselection mechanism with all path search algorithms enhances the start-of-the-art method SubgraphRAG on Qwen2.5-7b, as shown in Figure~\ref{fig:qwen_path_search}.
Employing Random Walk as path search algorithm shows a smaller improvement with the number of triples being 100 and 200.
This may be due to the limited number of walks, which fails to incorporate sufficient graph structure information.
The other two path search algorithms, BFS and Dijkstra, result in considerable improvement on CWQ, while Dijkstra spends less time for path searching.
For example, in the case of 50 triples, BFS can achieve a substantial improvement of 4.93\%, from 45.23 to 50.16, and Dijkstra can provide a similar increase of 4.9\%, from 45.23 to 50.13 on Hit@1.

\subsubsection{Different Pooling Strategies}
\begin{figure}[ht]
    \centering
    \includegraphics[width=0.5\textwidth]{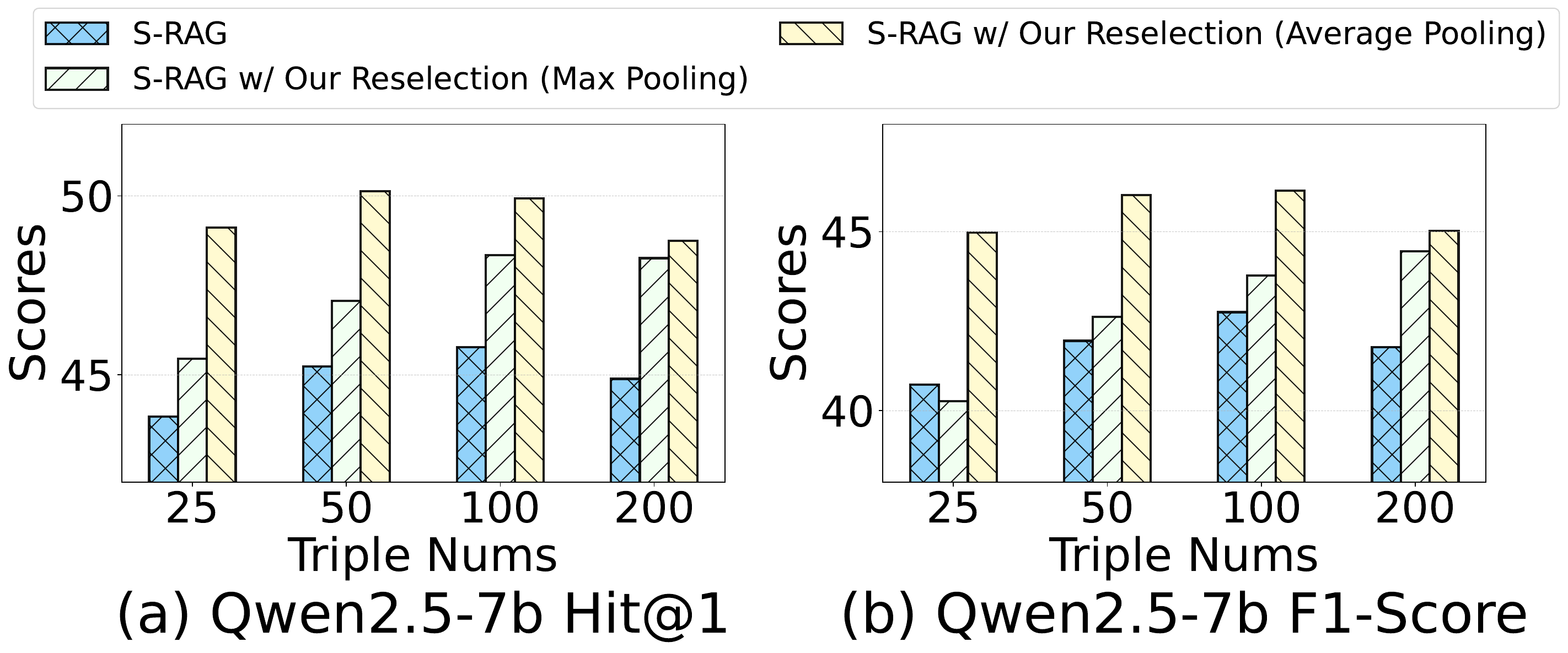}
    \caption{Reselection Results of Pooling Strategies on Qwen2.5-7b}
    \label{fig:qwen_pooling_strategy}
\end{figure} 

On Qwen2.5-7b, triple reselection mechanism with different pooling strategies still improves SubgraphRAG, as shown in Figure~\ref{fig:qwen_pooling_strategy}. 
Notably, SubgraphRAG with max pooling can achieve better results than the original SubgraphRAG across all triple numbers, both Hit@1 and F1-score.
Max pooling can provide a smoothing effect for triples along the path. However, since it retains only the maximum along that path, it may loss some important neighborhood information.
In contrast, average pooling demonstrates better improvement compared to max pooling across various parameter settings.
For example, max pooling provides a 1.84\% improvement over not using path pooling from 45.23 to 47.07, while average pooling brings 3.06\% more improvement than max pooling from 47.07 to 50.13 with 50 triples, demonstrating its significant effectiveness. 

\subsection{Positional Reranking on Llama3.1-8b}
\subsubsection{Different Path Search Algorithms}
\begin{figure}[ht]
    \centering
    \includegraphics[width=0.5\textwidth]{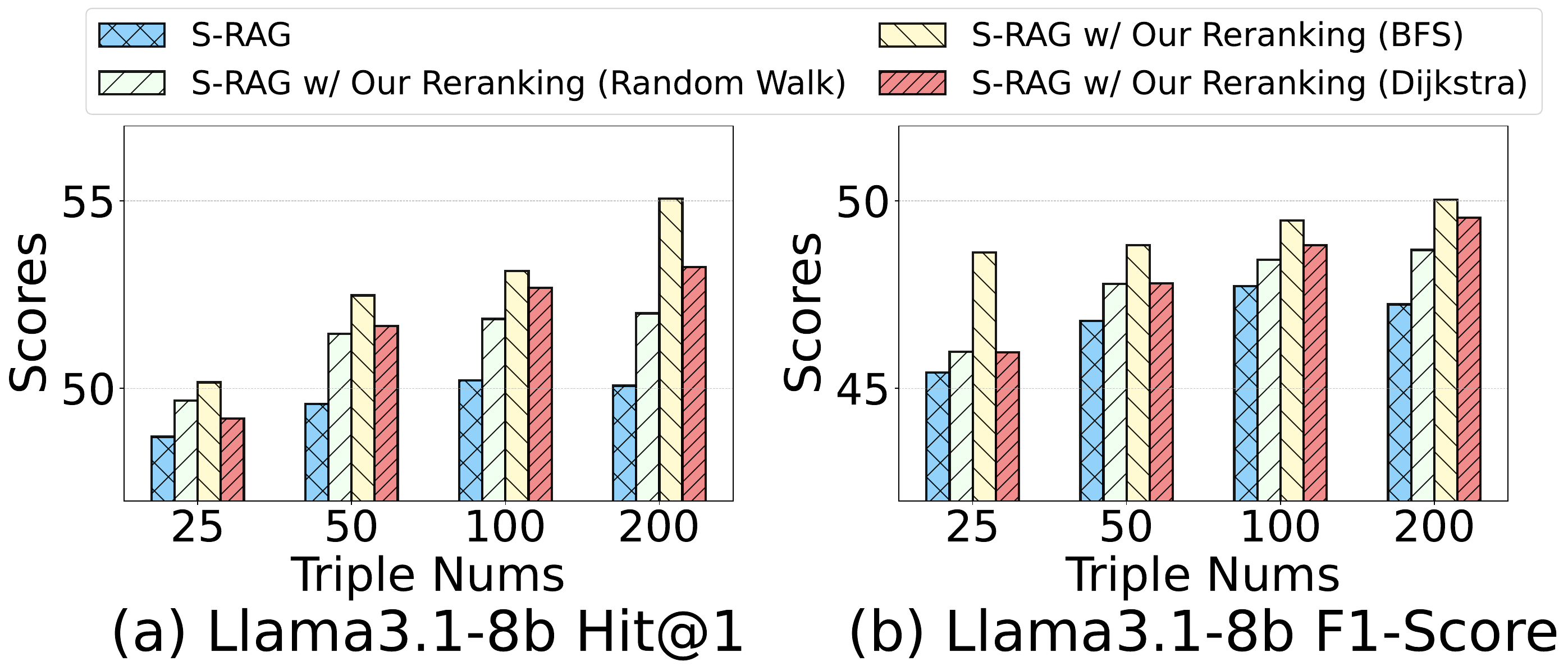}
    \caption{Reranking Results of Path Search Algorithms on Llama3.1-8b}
    \label{fig:path_search_reranking}
\end{figure}
Figure~\ref{fig:path_search_reranking} presents the performance of positional reranking mechanism with different path search algorithms on Llama3.1-8b. 
Overall, Random Walk, BFS and Dijkstra exhibit effectiveness in the knowledge graph question answering scenario.
Similar to the previous triple reselection mechanism, BFS delivers a better performance boost at the cost of longer path search times.
Since the number of paths searched by Random Walk remains consistent across different triple numbers, it performs better when the number of triples is low, while its performance does not significantly improve when the number of triples increase. This indirectly highlights the importance of incorporating sufficient structure information of KGs.

\subsubsection{Different Pooling Strategies}
\begin{figure}[ht]
    \centering
    \includegraphics[width=0.5\textwidth]{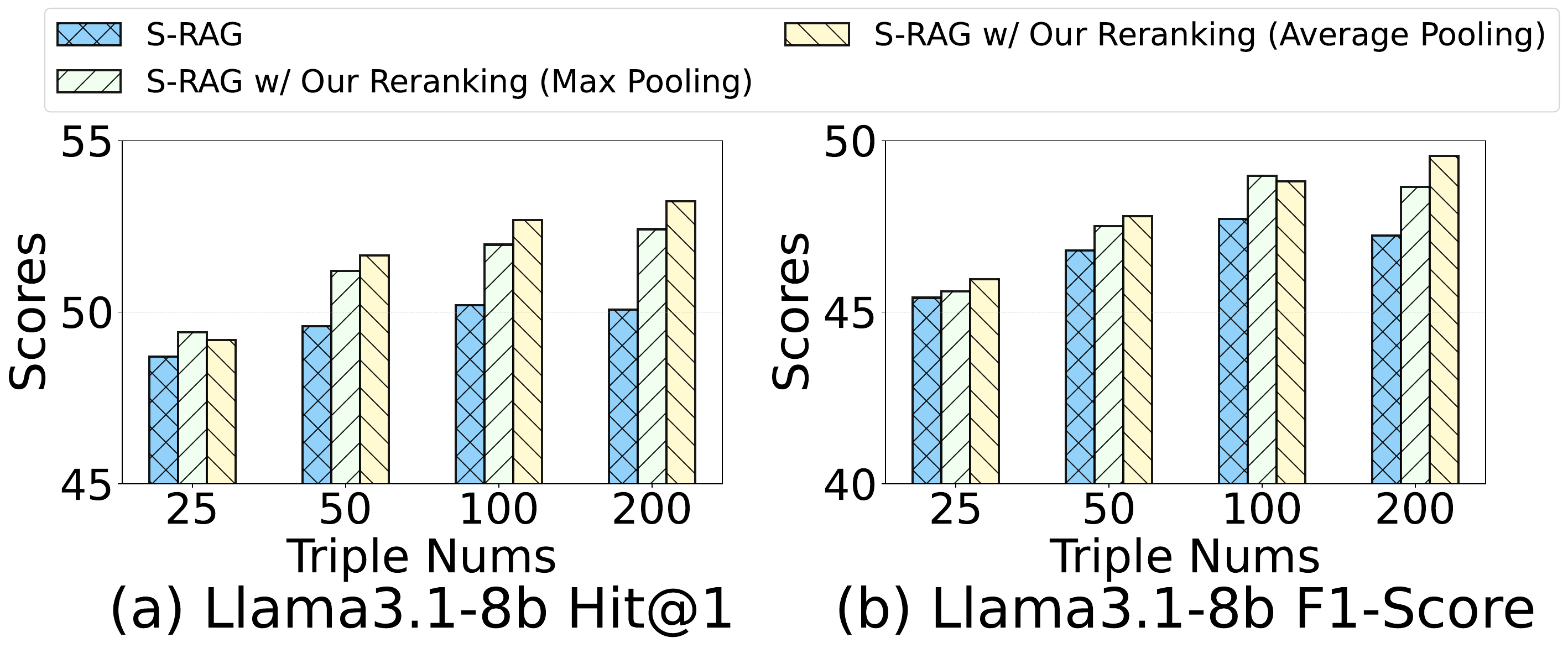}
    \caption{Reranking Results of Pooling Strategies on Llama3.1-8b}
    \label{fig:pooling_reranking}
\end{figure}
Figure~\ref{fig:pooling_reranking} presents the results of positional reranking mechanism with different pooling strategies on Llama3.1-8b. 
Both positional reranking with max pooling and positional reranking with average pooling show promising results. 
Notably, positional reranking with average pooling outperforms with max pooling in the majority of cases, whether in terms of Hit@1 or F1-Score.
For example, with 200 triples, position reranking mechanism with average pooling improves SubgraphRAG by 2\% on Hit@1 and by 1\% on F1-Score.

\subsection{Positional Reranking on Qwen2.5-7b}
\subsubsection{Different Path Search Algorithms}
\begin{figure}[ht]
    \centering
    \includegraphics[width=0.5\textwidth]{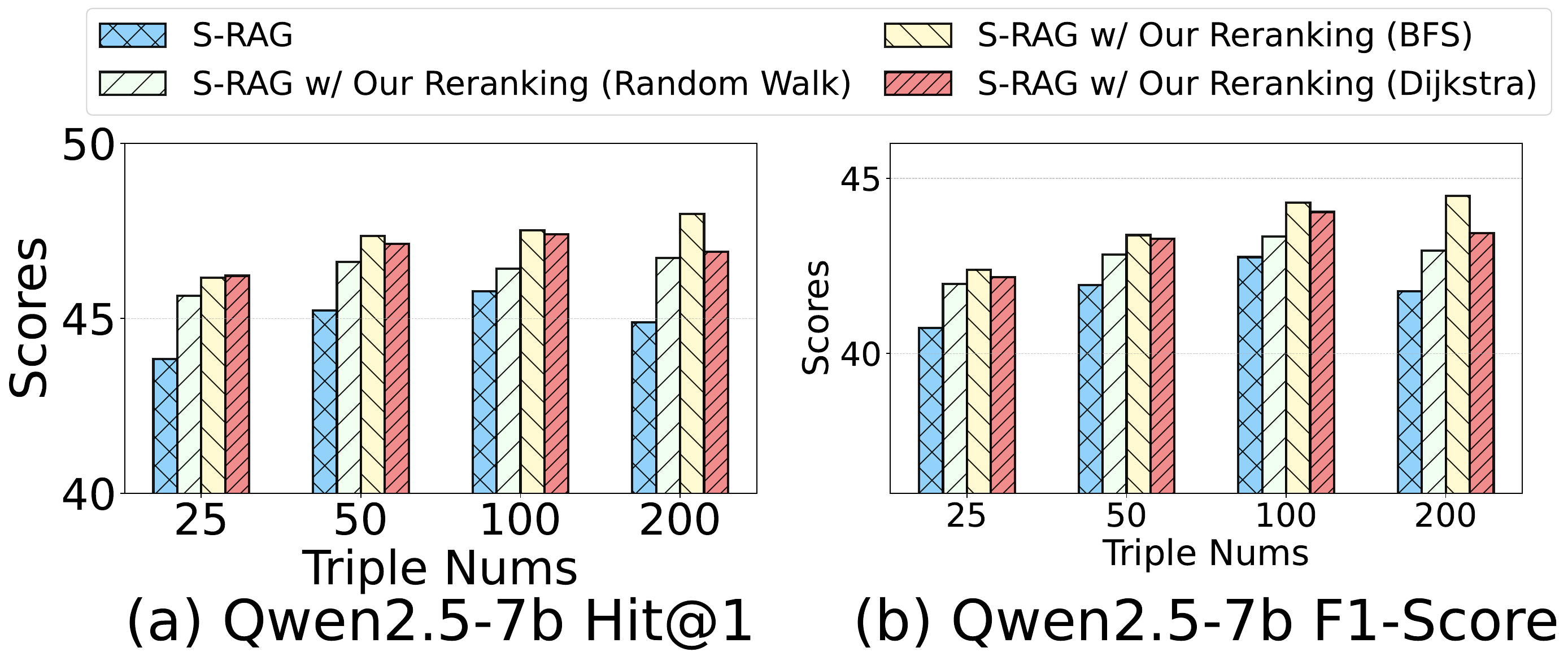}
    \caption{Reranking Results of Path Search Algorithms on Qwen2.5-7b}
    \label{fig:qwen_path_search_reranking}
\end{figure}
Figure~\ref{fig:qwen_path_search_reranking} summarizes Hit@1 and F1-Score of positional reranking mechanism with different path search algorithms on Qwen2.5-7b.
Consistent with previous results, all three path search algorithms are effective in enhancing the performance of SubgraphRAG.
As the number of triples increases, the Hit@1 and F1-Score of SubgraphRAG show a initially rising and then declining trend. 
However, when enhanced with a positional reranking mechanism based on BFS path search algorithm, both metrics demonstrate a consistently upward trend.
For instance, with 500 triples, the Hit@1 of SubgraphRAG is 45.77, while it is 47.52 with BFS, which brings about an increase of 1.75\%.
The result emphasizes the effectiveness of incorporating the structure information of KGs.

\subsubsection{Different Pooling Strategies}
\begin{figure}[ht]
    \centering
    \includegraphics[width=0.5\textwidth]{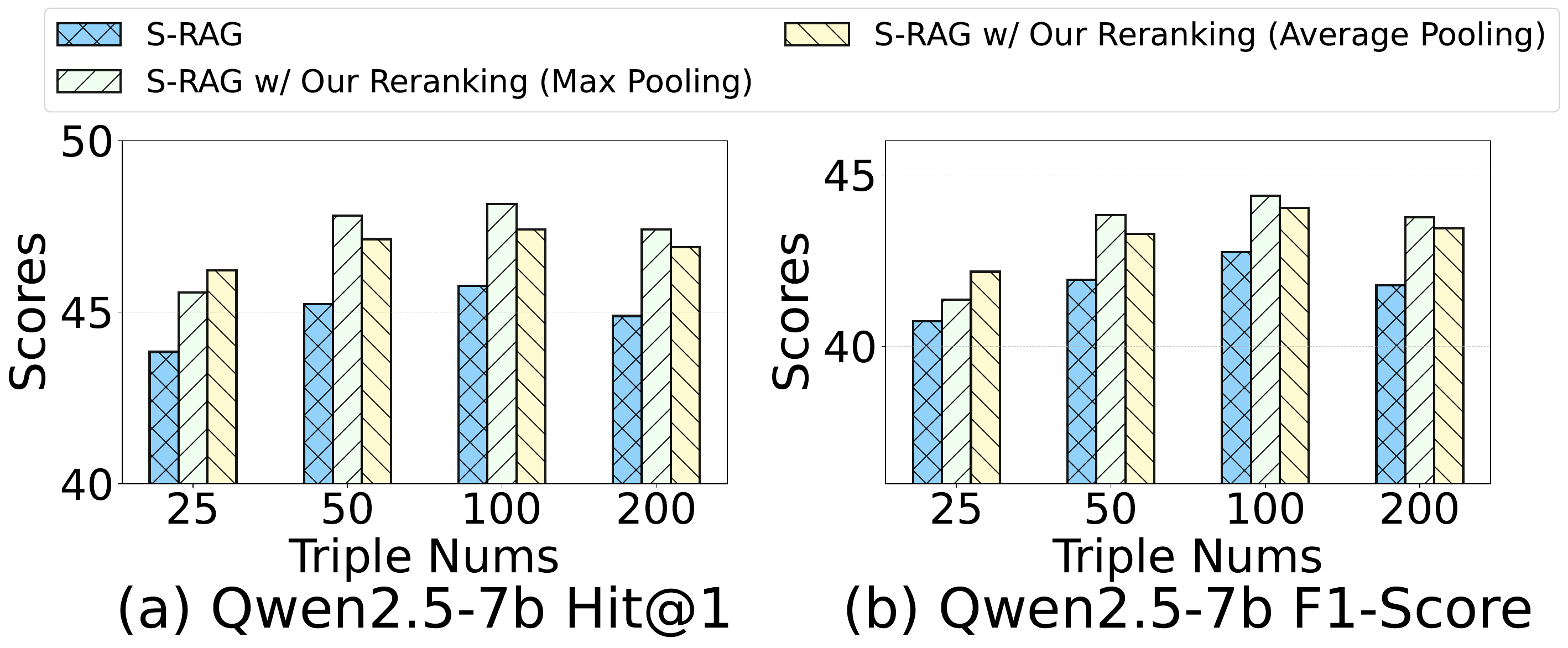}
    \caption{Reranking Results of Pooling Strategies on Qwen2.5-7b}
    \label{fig:qwen_pooling_reranking}
\end{figure}
As illustrated in Figure~\ref{fig:qwen_pooling_reranking}, positional reranking mechanism with different pooling strategies all enhanced SubgraphRAG on Qwen2.5-7b.
Similar benefits are observed when leveraging triple reselection mechanism and on Llama3.1-8b. 
As commonly used pooling strategies, both max pooling and average pooling demonstrate significant roles here.
However, surprisingly, when the number of triples increases from 50 to 200, the positional reranking mechanism using max pooling achieves better enhancement effect than using average pooling.
For example, with 50 triples, the Hit@1 of SubgraphRAG is 45.23, while it increases to 47.52 when using max pooling, resulting in an improvement of 2.23\%. 
In contrast, using average pooling only yields a 1.9\% increase.

\onecolumn
\subsection{Prompts}
\label{appendix:prompt}
The following is the detailed prompt template used in SubgraphRAG for all experiments.
Due to \textit{Recency Bias} that tokens appearing later in the sequence tend to receive higher attention weights, we place triples in ascending order
, aligned with all other experiments. 
\begin{tcolorbox}[title=SubgraphRAG Prompt]
	\noindent\textbf{System:} 
    
    Based on the triplets retrieved from a knowledge graph, please answer the question. Please return formatted answers as a list, each prefixed with "ans:".

    \noindent\rule{\textwidth}{0.7pt} 

    \noindent\textbf{User:} 

    \noindent\textit{Triplets:}\\
    (m.011zsc4\_, organization.leadership.organization, San Francisco Giants)\\
    (m.0crtd80, sports.sports\_league\_participation.league, National League West)\\
    \ldots\\
    (San Francisco Giants, time.participant.event, 2014 Major League Baseball season)\\
    (San Francisco Giants, time.participant.event, 2012 Major League Baseball season)\\
    (AT\&T Park, location.location.events, 2010 World Series)\\
    (San Francisco Giants, sports.professional\_sports\_team.owner\_s, Bill Neukom)\\
    (San Francisco Giants, time.participant.event, 2010 Major League Baseball season)\\
    (San Francisco Giants, sports.sports\_team.championships, 2010 World Series)\\
    (San Francisco Giants, time.participant.event, 2012 World Series)\\
    (Crazy Crab, sports.mascot.team, San Francisco Giants)\\
    (San Francisco Giants, time.participant.event, 2010 World Series)\\
    (San Francisco Giants, sports.sports\_team.championships, 2012 World Series)\\
    (San Francisco Giants, sports.sports\_team.team\_mascot, Crazy Crab)\\
    (San Francisco Giants, sports.sports\_team.championships, 2014 World Series)\\
    (Lou Seal, sports.mascot.team, San Francisco Giants)\\

    \noindent\textit{Question:}
    What year did the team with mascot named Lou Seal win the World Series?

    \noindent\rule{\textwidth}{0.7pt} 

    \noindent\textbf{Assistant:} 

    To find the year the team with mascot named Lou Seal won the World Series, we need to find the team with mascot named Lou Seal and then find the year they won the World Series. From the triplets, we can see that Lou Seal is the mascot of the San Francisco Giants. Now, we need to find the year the San Francisco Giants won the World Series. From the triplets, we can see that San Francisco Giants won the 2010 World Series and 2012 World Series and 2014 World Series. So, the team with mascot named Lou Seal (San Francisco Giants) won the World Series in 2010, 2012, and 2014. Therefore, the formatted answers are: 
    \\ans: 2014 World Series
    \\ans: 2012 World Series
    \\ans: 2010 World Series

    \noindent\rule{\textwidth}{0.7pt} 

    \noindent\textbf{User:} 

    Triplets:\\
    Question:\\
\end{tcolorbox}
\end{document}